\documentclass[runningheads]{llncs}
\usepackage[T1]{fontenc}
\usepackage{times}
\usepackage{soul}
\usepackage{url}
\usepackage[hidelinks]{hyperref}
\usepackage[utf8]{inputenc}
\usepackage[small]{caption}
\usepackage{graphicx}
\usepackage{amsmath}
\usepackage{amssymb}
\usepackage{booktabs}
\usepackage{algorithmic}
\usepackage[switch]{lineno}
\usepackage{comment}
\usepackage{subcaption}
\usepackage[ruled,vlined,linesnumbered]{algorithm2e}
\usepackage{color}

\begin{document}
\title{Unassigned Agents in Compilation-based\\Multi-agent Path Finding}

\author{Pavel Surynek
}

%\author{First Author\inst{1}\orcidID{0000-1111-2222-3333} \and
%Second Author\inst{2,3}\orcidID{1111-2222-3333-4444} \and
%Third Author\inst{3}\orcidID{2222--3333-4444-5555}}

\authorrunning{P. Surynek}
\institute{Faculty of Information Technology, Czech Technical University in Prague\\Th\'{a}kurova 9, 163 00, Prague, Czechia\\
\email{pavel.surynek@fit.cvut.cz}
}
\maketitle              % typeset the header of the contribution
\begin{abstract}
Compilation-based techniques represent an important stream of solvers for multi-agent path finding (MAPF) due to their modularity and adaptability for non-standard variants of the problem. While in the standard MAPF the task is to navigate all agents from their initial positions to given individual goal positions without any collision, variants where a different requirement for agents is used are also relevant. Such a variant is MAPF with unassigned agents (UA-MAPF) where some agents have the same setting as in the standard MAPF with initial positions and goals while the remaining agents have the initial position but have no goal - unassigned agents. Despite unassigned agent do not need to reach any goal position they have to be moved out of the way of the standard agents if needed which represent a specific challenge. We show in this paper that UA-MAPF can be expressed in recent compilation-based techniques for MAPF based on formulating the problem as Boolean satisfiability, namely we adapt SMT-CBS and NRF-SAT, the recent solvers based on counterexample guided abstraction refinement and non-refined abstractions.

\end{abstract}
\section{Introduction}
Multi-agent path finding (MAPF) \cite{DBLP:journals/jair/Ryan08,DBLP:conf/icra/Surynek09,DBLP:journals/jair/WangB11} is a task of finding non-conflicting paths for $m \in \mathbb{N}$ agents $A=\{a_1,a_2,...,a_m\}$ that move in an undirected graph $G=(V,E)$ across its edges such that each agent reaches its goal vertex from the given start vertex via its path. Starting configuration of agents is defined by a simple assignment $s: A \rightarrow V$ and the goal configuration is defined by a simple assignment $g: A \rightarrow V$. A conflict between agents is usually defined as simultaneous occupancy of the same vertex by two or more agents or as a traversal of an edge by agents in opposite directions.

MAPF with unassigned agents (UA-MAPF) \cite{DBLP:conf/aaai/FelnerS26} generalizes MAPF via introducing a set of unassigned agents $U=\{u_1,u_2,...,u_{m'}\}$. Starting configuration of agent in $G$ is naturally extended for unassigned agents: $s: \{A \cup U\} \rightarrow V$, however the goal configurations remains defined only for the standard (assigned) agents: $g: A \rightarrow V$.

\begin{figure}[t]
    \centering
    \includegraphics[trim={4.0cm 22.8cm 3.5cm 3.0cm},clip,width=0.9\textwidth]{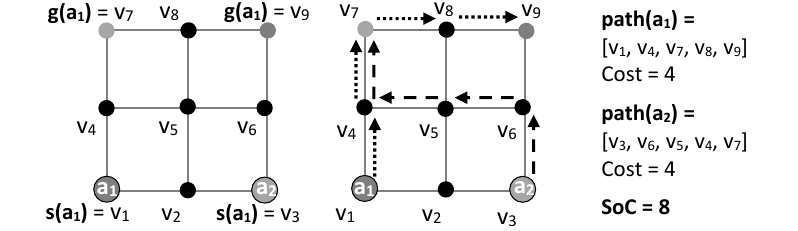}
    %\vspace{-0.8cm}
    \caption{Multi-agent path finding (MAPF) with agents $a_1$ and $a_2$.}
    \label{figure-MAPF}
\end{figure}

Many problems in robotics, urban traffic optimization, circuit design, and computer games can be regarded from the perspective of MAPF as listed by various surveys including \cite{DBLP:journals/corr/0001KA0HKUXTS17}. UA-MAPF extends the applicability to other practical cases. Various motivations are listed for UA-MAPF in \cite{DBLP:conf/aaai/FelnerS26}. Let us  name few examples where UA-MAPF is relevant: automated parking lots where we need to put cars out of the way for a car leaving the parking lot, automated warehouses where some robots are not working but need to get out of the way of robots currently working, even packages themselves in warehouses can be treated as unassigned agents that need to make room for agents.

Particular aspect that is motivated by practice and makes MAPF challenging is the need to find the optimal solutions with respect to some cumulative cost \cite{DBLP:conf/aaai/YuL13}. Commonly used cumulative costs in MAPF are {\em makespan} and {\em sum-of-costs}\cite{DBLP:conf/raai/Stern19}. The makespan corresponds to the length of the longest agent's path. The sum-of-costs is the sum of costs of individual paths which corresponds to the sum of unit costs of actions, the wait actions including. An example of MAPF problem and its sum-of-costs optimal solution is shown in Figure \ref{figure-MAPF}.

Cumulative objectives are also important in UA-MAPF, however the presence of unassigned agents allows for more combinations than the standard MAPF. The following objectives are defined for UA-MAPF in \cite{DBLP:conf/aaai/FelnerS26}:

\begin{itemize}
\item {\bf Sum of Service Time (SST)} - this cost is the sum of time steps required to move each assigned agent to its target. SST corresponds maximizing the system throughput.
\item {\bf FUEL} - this cost is the total number of moves made by all the agents (and waiting does not cost). This cost corresponds to minimizing the energy costs required to move the agents.
\item {\bf Number of unassigned agents that move (NUA)} - here we do not care about the lengths of the paths but aim to
minimize the number of unassigned agents that move.
\end{itemize}

Additional objectives or the combination of previous can be introduced. We introduce an objective {\bf FUEL+} that adds cost for waiting before the agent's final move (the motivation for this cost is that waiting still consumes fuel as the engine is on). This objective corresponds to the sum-of-cost objective from the standard MAPF.

\begin{figure}[h]
    \centering
    \includegraphics[trim={4.0cm 21.5cm 3.5cm 3.0cm},clip,width=0.9\textwidth]{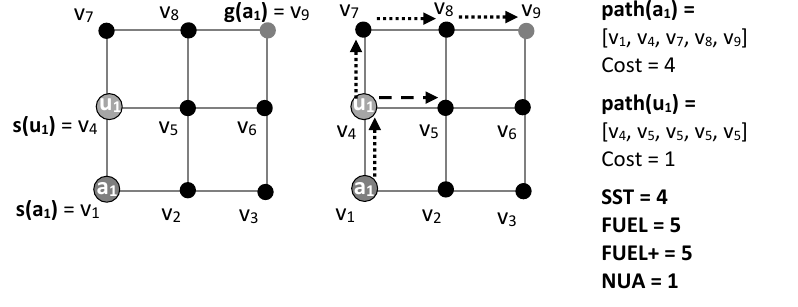}
    %\vspace{-0.8cm}
    \caption{Multi-agent path finding with unassigned agents (UA-MAPF) where $A=\{a_1\}$ and $U=\{u_1\}$. Unassigned agent $u_1$ gets out of the way of standard (assigned) agent $a_1$. Then agent $u_1$ does not need to go anywhere so it waits in $v_5$.}
    \label{figure-UA-MAPF}
\end{figure}

An example of UA-MAPF is shown in Figure \ref{figure-UA-MAPF}, values of abovementioned objectives are shown in the figure (the solution is optimal in SST, but sub-optimal with respect to FUEL, FUAL+, and NUA (alternative path $[v_1,v_2,v_3,v_6,v_9]$ for $a_1$ exists).

%NRF-SAT, NRF-CBS, NRF-MAPF, NCEGAR-MAPF, SMT-CBS, NRF-SAT, NRF-CEGAR-MAPF

We will address UA-MAPF in this paper from the perspective of compilation-based techniques \cite{DBLP:conf/ijcai/Surynek22} that represent a major alternative to search-based solvers \cite{DBLP:journals/ai/SharonSGF13,DBLP:journals/ai/SharonSFS15} for MAPF. Compilation-based solvers reduce the input MAPF instance to an instance in a different well established formalism for which an efficient solver exists. Such formalisms are for example {\em constraint programming} (CSP) \cite{DBLP:books/daglib/0016622}, {\em Boolean satisfiability} (SAT) \cite{biere2021handbook}, or {\em mixed integer linear programming} (MILP) \cite{rader2010deterministic}.

\subsection{Related Work}

\begin{figure}[h]
    \centering
    \includegraphics[trim={2.1cm 22.5cm 3.5cm 3.0cm},clip,width=0.9\textwidth]{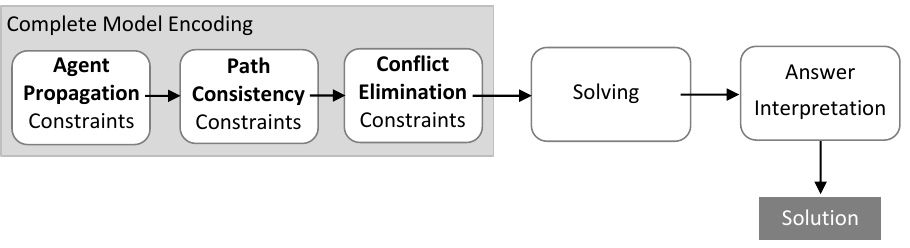}
    %\vspace{-0.8cm}
    \caption{Schematic diagram of the basic MAPF compilation with complete model.}
    \label{figure-COMPILATION}
\end{figure}

The basic compilation scheme for the sum-of-costs optimal MAPF solving has been introduced by the MDD-SAT solver \cite{DBLP:conf/ecai/SurynekFSB16} that uses so called {\em complete models} to compile MAPF instances into SAT (see Figure \ref{figure-COMPILATION}). The target Boolean formula of the complete model is satisfiable if and only if the input MAPF has a solution of a specified sum-of-costs. The complete model as introduced in MDD-SAT consists of three group of constraints:

\begin{itemize}
\item {\bf Agent propagation} constraints - these constraints ensure that if an agent appears in vertex $v$ at time step $t$ then the agent appears in some neighbor of $v$ (including $v$ itself) at time step $t+1$. The side effect of these constraints is that the agent never disappears. Cost calculation and bounding is done together with agent propagation.
\item {\bf Path consistency} constraints - these constrains ensure that agents move along proper paths, that is, agents do not duplicate themselves and do not appear spontaneously.
\item {\bf Conflict elimination} constraints - to ensure that agents do not conflict with each other according to the MAPF rules (vertex and edge conflicts).
\end{itemize}

Compilation-based techniques experienced significant progress recently via introduction of {\bf laziness} and so called {\em incomplete models}. In lazy compilation schemes, conflict elimination constraints are omitted at the beginning and the (incomplete) model is initially solved without them. This leads to potential conflicts that are iterativelly eliminated by refinements of the model with conflict elimination constraints. The advantage is that the model can end-up with significantly fewer constraints than the complete model. The representatives of incomplete models for various target formalisms include the CSP-based LazyCBS \cite{DBLP:conf/aips/GangeHS19}, SAT-based SMT-CBS \cite{DBLP:conf/ijcai/Surynek19}, and MILP-based BCP \cite{DBLP:conf/ijcai/LamBHS19}.

\begin{figure}[h]
    \centering
    \includegraphics[trim={2.7cm 20.2cm 2.8cm 2.9cm},clip,width=0.9\textwidth]{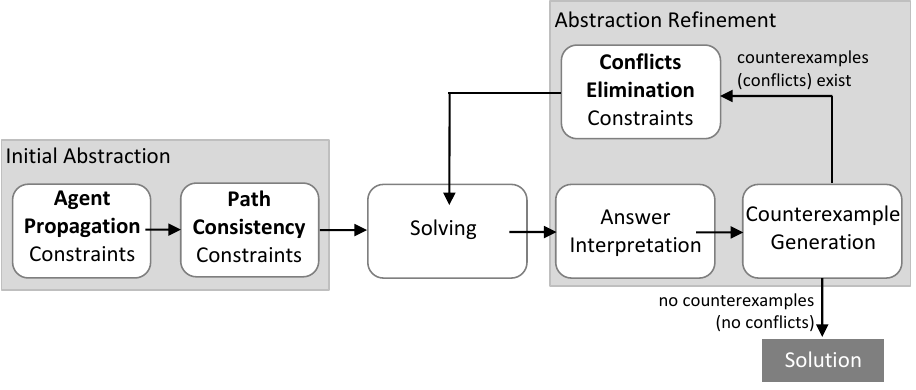}
    %\vspace{-0.8cm}
    \caption{Schematic diagram of counterexample guided abstraction refinement (CEGAR) for MAPF. Conflicts between agents are treated as counterexamples and eliminated in the abstraction refinement loop. }
    \label{figure-CEGAR}
\end{figure}

The idea of laziness is pushed further in \cite{DBLP:conf/ictai/Surynek23}, where the NRF-SAT solver is introduced. NRF-SAT builds on the idea of {\em counterexample guided abstraction refinement} (CEGAR) \cite{DBLP:conf/cav/ClarkeGJLV00} (see Figure \ref{figure-CEGAR}). An abstraction in the context of compilation for MAPF can be understood as constraint omission. NRF-SAT further enhances the CEGAR perspective of MAPF with so called {\bf non-refined abstractions}. While the standard CEGAR corresponds to incomplete models where abstractions and refinements are done with respect to conflicts, non-refined abstractions leave certain abstractions intentionally non-refined, that is, certain constraints are intentionally never added to the model.

\begin{figure}[h]
    \centering
    \includegraphics[trim={2.4cm 20.2cm 2.4cm 3.0cm},clip,width=0.9\textwidth]{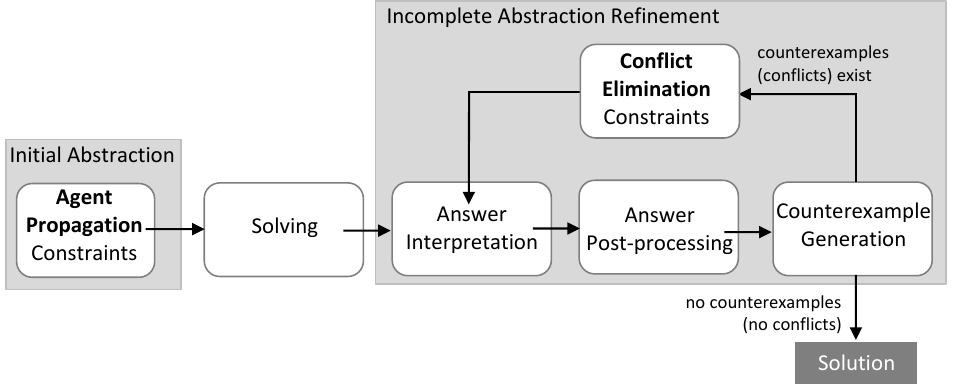}
    %\vspace{-0.8cm}
    \caption{Schematic diagram of a CEGAR problem solver for MAPF with non-refined abstractions.}
    \label{figure-NRF-CEGAR}
\end{figure}

NRF-SAT introduces non-refinement with respect to path-consistency constraints. Absence of path-consistency constrains means that agents can duplicate themselves or can appear spontaneously at individual time steps of the resulting plan. A special {\bf post-processing} step is therefore necessary that extracts a valid path for each agent from the resulting plan - see Figure \ref{figure-NRF-CEGAR}. An interpretation offered in \cite{DBLP:conf/ictai/Surynek23} suggests that the model with path consistency constraints omitted corresponds to finding trees or directed acyclic graphs (DAG) that connects agents' initial positions and goals instead of paths. A valid path can be extracted from a tree or from a DAG by a polynomial time post-processing step.

\subsection{Contribution}

The contribution of this paper is an adaptation of the SMT-CBS and NRF-SAT solvers for UA-MAPF. This demonstrates adaptability and modularity of compilation-based approaches in general and specifically those based on SAT. Our experimental evaluation shows important insights about UA-MAPF. The most important observation is counterintuitive impact of relaxations intruduced by unassigned agents.

\section{Time Expansion for Unassigned Agents}

The crucial technique for reducing MAPF to SAT is {\em time expansion} using {\em time expanded graphs} (TEGs) and {\em multi-valued decision diagrams} (MDDs) introduced in \cite{DBLP:conf/ecai/SurynekFSB16}. TEG is a structure that represents positions of agents at individual time steps. For a given makespan $\mu$ and graph $G=(V,E)$ with $V=\{v_1,v_2,...,v_n\}$, a TEG with respect to $\mu$ and $G$, or TEG$(G,\mu)$ is a directed acyclic graph (DAG) that contains a copy of vertices of $G$ for each individual time steps $t=0,1,...,\mu$. That is, vertices $v_i^t$ for $v_i \in V$ are introduced. For each undirected edge $\{v_i, v_j\} \in E$, directed edges $(v_i^t,v_j^{t+1})$ and $(v_j^t,v_i^{t+1})$ are introduced into TEG$(G,\mu)$ for each $t=0,1,...,{\mu}-1$. In addition to this, edges $(v_i^t,v_i^{t+1})$ are added to TEG$(G,\mu)$ to represent wait actions of agents.

\begin{figure}[t]
    \centering
    \includegraphics[trim={2.8cm 21.8cm 3.0cm 2.5cm},clip,width=1.0\textwidth]{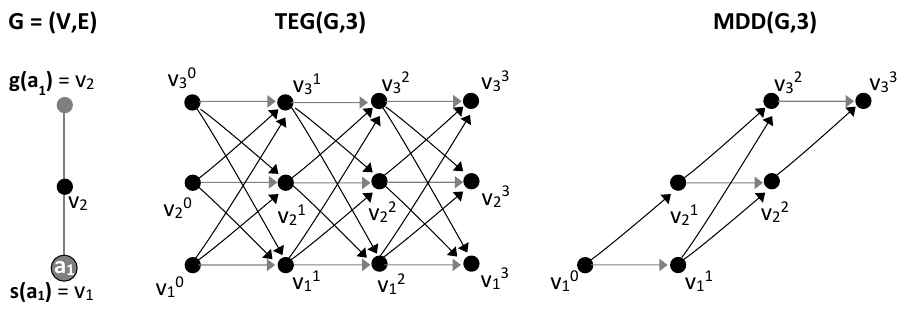}
    %\vspace{-0.8cm}
    \caption{An example of TEG and MDD for makespan $\mu = 3$ for a graph consisting of three vertices.}
    \label{figure-TEG-MDD}
\end{figure}

TEGs are intuitive tools to establish a correspondence between solutions of MAPF and directed paths in TEGs. A directed path in   TEG$(G,\mu)$ connecting $v_{s(a_k)}^0$ and $v_{g(a_k)}^{\mu}$ represents a plan for agent $a_k$ of makespan $\mu$. Let us note that unlike a path in the original $G$, TEG$(G,\mu)$ allows for multiple visits of the same vertex by an agent thanks to the time expansion. The solution of MAPF coordinating multiple paths in non-conflicting way can be represented in TEG$(G,\mu)$ as multiple directed paths that are vertex and edge disjoint which reflects vertex and edge conflicts, alternatively multiple TEGs can be used, one per agent. An example of TEG is shown in Figure \ref{figure-TEG-MDD}. A more detailed description can be found in \cite{DBLP:conf/ecai/SurynekFSB16}.

TEG$(G,\mu)$ can be directly used to build a Boolean formula that is satisfiable if and only if a solution of a corresponding MAPF exists. The core idea is to introduce a Boolean variable $\mathcal{X}_{i,k}^t$ for each vertex of TEG$(G,\mu)$ and agent $a_k \in A$. The interpretation of $\mathcal{X}_{i,k}^t$ is that it is $\mathit{TRUE}$ if and only if agent $a_k$ visits vertex $v_i$ at time step $t$. Boolean variables for edges can be used as well as it helps to express the conflict avoidance constraints and cost functions. We list a constraint for vertex conflict for illustration, for the full list we refer the reader to \cite{DBLP:conf/ecai/SurynekFSB16}:

\begin{equation}
  \neg \mathcal{X}_{i,k}^t \wedge \neg \mathcal{X}_{i,k'}^t. \\
  \label{eq-conflict}
\end{equation}

Constraint \ref{eq-conflict} forbids simultaneous occurrence of agents $a_k$ and $a_{k'}$ in vertex $v_i$ at time step $t$.
TEGs can be understood as a pedagogical aid as they are too inefficient for practical use. This is because TEGs represent {\bf unreachable} vertices, so corresponding Boolean variables can never be set to $\mathit{TRUE}$ and therefore makes the resulting Boolean formula unnecessarily large. Consider for example time step 1 in TEG$(G,\mu)$ for agent $a_k$. Only vertices that are one-step reachable from $v_{s(a_k)}^0$ are relevant, that is immediate neighbors of $v_{s(a_k)}^0$ in TEG$(G,\mu)$ are relevant for introducing Boolean variables to represent presence of $a_k$ at time step 1. Similarly for vertices at time step $\mu-1$ with respect to the goal vertex $v_{g(a_k)}^{\mu}$. Again only one-step reachable vertices from $v_{g(a_k)}^{\mu}$ are relevant for introducing Boolean variables to represent presence of $a_k$ at time step $\mu - 1$. In general, vertex $v_i^t$ is relevant if:

\begin{equation}
   \mathit{dist}(s(a_k),v_i) \leq t \wedge \mathit{dist}(v_i,g(a_k)) \leq \mu - t, \\
   \label{eq-MDD}
 \end{equation}

where $\mathit{dist}$ denotes the {\em shortest path distance}. Condition \ref{eq-MDD} is a basis for MDDs that can be regarded as filtered TEGs. An example of MDD is shown in Figure \ref{figure-TEG-MDD}. MDDs results in a significantly smaller formula. Constraints to represent {\bf agent propagation}, {\bf path consistency}, and {\bf conflict elimination} can be introduced on top of Boolean variables  $\mathcal{X}_{i,k}^t$ derived from MDDs for individual agents instead of TEGs.

All the advanced solvers based on lazy encodings and counterexample guided abstraction refinements such as SMT-CBS \cite{DBLP:conf/ijcai/Surynek19} and NRF-SAT \cite{DBLP:conf/ictai/Surynek23} are based on the underlying encoding derived from MDDs.

\subsection{Unassigned Agents in MDDs}

MDD-SAT as well as other SAT-based algorithms iterates makespan $\mu$ from the lower bound $\mu_0$ determined as: $\max_{k \in \{1,2,...,m\}}{\mathit{dist}(s(a_k),g(a_k))}$ (that is the length of longest path connecting agents' starting position with the goal via shortest path) until a solution is found. For each makespan $\mu$ a formula according to the above scheme is constructed and consulted with the SAT solver \cite{DBLP:journals/ijait/AudemardS18}. In the case of positive answer, a solution is extracted from the truth assignment, otherwise the search continues with makespan $\mu + 1$. The algorithm is {\em makespan optimal} as it searches for the first solvable makespan and {\em semi-complete} that is, it terminates for solvable instances but not for unsolvable ones \footnote{Unsolvable cases can be treated by running first a polynomial time complete sub-optimal algorithm such as Push-and-Rotate \cite{DBLP:journals/jair/WildeMW14}.}.

\begin{figure}[t]
    \centering
    \includegraphics[trim={2.8cm 21.8cm 8.0cm 2.5cm},clip,width=0.6\textwidth]{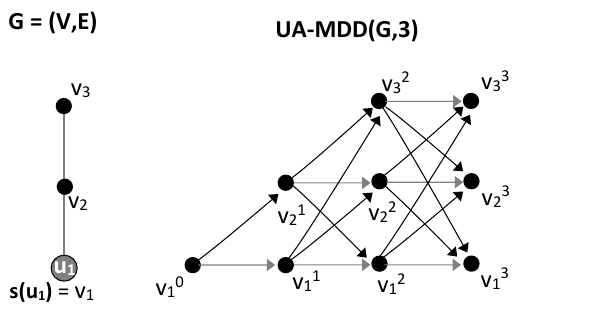}
    %\vspace{-0.8cm}
    \caption{An example of UA-MDD for makespan $\mu = 3$ for a graph consisting of three vertices.}
    \label{figure-UA-MDD}
\end{figure}

Optimality with respect to other cumulative costs is established via additional bound introduced over the edge variables (each edge variable corresponds to agent's action). For the FUEL+ cost, we augment the above incremental scheme with an additional bound $f^+$. We start with the lower bound $f^+_0$ for $f^+$ determined as: $\sum_{k = 1}^m{\mathit{dist}(s(a_k),}$ ${g(a_k))}$, this is the lower bound for FUEL+ (unassigned agents are initially assumed to wait only). The number of edge variables that can be set to $\mathit{TRUE}$ is bounded by $f^+$. Both $\mu$ and $f^+$ are incremented after a negative answer from the SAT solver. The algorithm finds a FUEL+ optimal solution. It is important to note, that incrementing both $\mu$ and $f^+$ allows for all extra FUEL+ above the lower bound $f^+_0$ to be consumed by a single agent. Let us also note that MDD contains edges for wait actions so FUEL+ consumption for waiting is reflected in the formula. This is analogous to the incremental scheme for the sum-of-costs objective as introduced in \cite{DBLP:conf/ecai/SurynekFSB16}.

Unassigned agents represent a difficulty as they are not required to arrive at certain goals. This allows unassigned agents to move freely. The free movement of unassigned agents cannot be easily restricted without affecting soundness of the solving process (for example unassigned agent blocking an assigned agent in a corridor must travel a long distance out of the corridor to free a way for the assigned agent). Directing the standard assigned agents towards their goals is important for reducing the size of MDD via removing unreachable vertices with respect goal vertices as done in equation (\ref{eq-MDD}). This size reduction is however not possible for unassigned agents as they have no goals. Removing unreachable vertices with respect to starting vertices is still possible for unassigned agents. The variant of MDD with unassigned agents will be called UA-MDD (for UA-MDD only the first part of equation (\ref{eq-MDD}) is applicable, Figure \ref{figure-UA-MDD}).
%An example of UA-MDD is shown in Figure \ref{figure-UA-MDD}.

\section{Experimental Evaluation}

We implemented support for unassigned agents in the SMT-CBS and NRF-SAT solvers. Both solvers and unassigned agent support are implemented in C++. The Glucose SAT solver \cite{DBLP:journals/ijait/AudemardS18} is used as the underlying solver. To support reproducibility of results we provide all source codes and experimental data on \texttt{github.com/surynek/boOX}. All experiments have been performed on a system with CPU AMD Ryzen 7 2700 3.2GHz, 32GB RAM, running Kubuntu Linux 24.

\begin{figure}[t]
    \centering
    \includegraphics[trim={2.0cm 17.0cm 2.5cm 2.5cm},clip,width=1.0\textwidth]{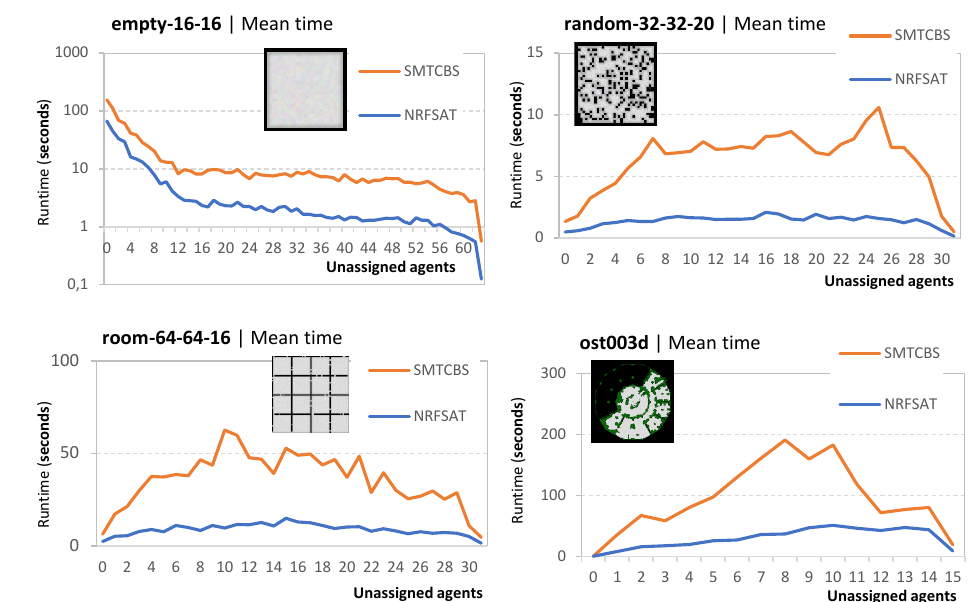}
    %\vspace{-0.8cm}
    \caption{Runtime results of SMT-CBS and NRF-SAT in {\bf FUEL+ optimal} UA-MAPF solving for the increasing number of unassigned agents in four different maps from the movingai.com benchmark collection: \texttt{empty-16-16}, \texttt{random-32-32-20},\texttt{room-64-64-16}, and \texttt{ost003d}. The plot for \texttt{empty-16-16} uses a logarithmic scale for runtime, the plots of other maps use a linear scale. Each data point is the mean value of 25 random measurements.}
    \label{figure-runtime}
\end{figure}

\subsection{Impact of Agent Unassignment}
 
 It is hard to predict what will be the impact of introducing unassigned agents in SAT-based MAPF approaches as it has two contradictory effects. On one hand, unassigned agents represent a relaxation of the problem. On the other hand, unassigned agents induce UA-MDDs that are larger than MDDs for corresponding agents.

We measured runtime for the increasing number of unassigned agents on structurally different maps from the movingai.com benchmark collection \cite{DBLP:journals/tciaig/Sturtevant12}. We present part of the results concerning four maps: \texttt{empty-16-16} - a small baseline benchmark, \texttt{random-32-32-20} - a map with random obstacles, \texttt{room-64-64-16} - a map with many bottlenecks connecting separate rooms, and \texttt{ost003d} - a large map from a computer game. The number of unassigned agents has been increased from 0 to 1 less than the total number of agents (the total number of agents ranged from 16 to 64). Runtime for 25 instances per number of agents with different settings of starting and goal positions was measured. The time limit for the solver was set to 5 minutes.

Runtime results for SMT-CBS and NRF-SAT are reported in Figure \ref{figure-runtime}, the mean runtime out of the 25 instances per number of agents is reported. Results across all tested maps indicate that both hypothesized effects are manifested. For the initial increase in the number of unassigned agents, the UA-MAPF problem becomes more complicated for both solvers, evidently due to the increasing number of large UA-MDDs. Relaxation due to the greater freedom of unassigned agents is not significant  enough to manifest itself at the beginning. The relaxation effect prevails only for a large number of unassigned agents. These results can be observed for three larger maps:  \texttt{random-32-} \texttt{32-20}, \texttt{room-64-64-16}, and \texttt{ost003d}. The smallest map \texttt{empty-16-16} shows a different trend, the effect of relaxation prevails from the beginning which supports the explanation that large UA-MDDs affect runtime negatively since in \texttt{empty-16-16} the large size of UA-MDD has a negligible effect.

The hypothesized effects are more pronounced in SMT-CBS while in NRF-SAT also observable but are less pronounced. This is due to overall higher efficiency of the NRF-SAT solver that mitigates the large size of the formula via non-refined abstractions.

\section{Conclusion}

We study a recently introduced variant of multi-agent path finding (MAPF) with unassigned agents (UA-MAPF). Specifically we show how to adapt SAT-based solvers for UA-MAPF, two solvers SMT-CBS \cite{DBLP:conf/ijcai/Surynek19} and NRF-SAT \cite{DBLP:conf/ictai/Surynek23} were adapted for UA-MAPF. It was only necessary to change one of the solvers' components, namely the generation of UA-MDD instead of MDD, which demonstrates the modularity and adaptability of SAT-based approaches. The experimental evaluation shows a counterintuitive trend in larger maps where increasing number of unassigned agents causes longer runtimes while the intuitive expectation is that unassigned agents will relax the problem. The relaxation eventually prevails with majority of unassigned agents.

In future work, we would like to introduce unassigned agents also for continuous variants of the MAPF problem \cite{DBLP:journals/ai/AndreychukYSAS22}.

\bibliographystyle{splncs04}
\bibliography{references}

\end{document}